\documentclass[lettersize,journal]{IEEEtran}
\usepackage{amsmath,amsfonts}
\usepackage{algorithmic}
\usepackage{array}
\usepackage[caption=false,font=normalsize,labelfont=sf,textfont=sf]{subfig}
\usepackage{textcomp}
\usepackage{stfloats}
\usepackage{url}
\usepackage{verbatim}
\usepackage{graphicx}
\usepackage{threeparttable}
\usepackage{booktabs,tabularx}
\usepackage{multirow}
\usepackage{makecell}
\def\BibTeX{{\rm B\kern-.05em{\sc i\kern-.025em b}\kern-.08em
    T\kern-.1667em\lower.7ex\hbox{E}\kern-.125emX}}
\usepackage{balance}
\graphicspath{{./figs/}}

\begin{document}
\title{How to Use the IEEEtran \LaTeX \ Templates}

\title{Optimize-at-Capture: Highly-adaptive Exposure Controlling for In-Vehicle Non-contact Heart-rate Monitoring}
\author{Jieying Wang, Xinqi Cai, Caifeng Shan* \IEEEmembership{Senior Member, IEEE}, and Wenjin Wang*
	\thanks{This work is supported by the National Natural Science Foundation of China (62501366), Shandong Provincial Natural Science Foundation (ZR2024QF268), Shenzhen Key Industrial R\&D Program (ZDCYKCX20250901092803004), Shenzhen Medical Research Fund (D2402011), Shenzhen Science and Technology Program (JSGGKQTD20221103174704003), and Guangdong S\&T Program (2023ZT10Z002).}
	\thanks{This work involved human subjects in its research. Approval of all ethical and experimental procedures and protocols was granted by institutional review board of Southern University of Science and Technology under Application No. 20240150.}
	\thanks{Jieying Wang is with the College of Computer Science and Engineering, Shandong University of Science and Technology, Qingdao 266590, China (e-mail: jieying.wang@sdust.edu.cn)}
	\thanks{Wenjin Wang and Xinqi Cai are with the Department of Biomedical Engineering, College of Engineering, Southern University of Science and Technology, Shenzhen 518000, China (e-mail: wangwj3@sustech.edu.cn; 12411558@mail.sustech.edu.cn).}
	\thanks{Caifeng Shan is with the State Key Laboratory for Novel Software Technology and School of Intelligence Science and Technology, Nanjing University, Nanjing 210023, China (e-mail: cfshan@nju.edu.cn)}
	\thanks{* The corresponding authors.}
}

\markboth{Journal of \LaTeX\ Class Files,~Vol.~18, No.~9, September~2020}
{Jieying Wang \MakeLowercase{\textit{et al.}}: In-Vehicle Non-contact Heart-rate Monitoring}

\maketitle

\begin{abstract}
	Remote photoplethysmography (rPPG) holds great promise for continuous heart-rate monitoring of drivers in intelligent vehicles. However, its performance is severely degraded by the highly dynamic illumination changes. A critical yet overlooked factor is the lack of exposure controlling during video acquisition—most existing systems rely on either fixed exposure settings or camera build-in auto-exposure, both of which fail to maintain stable facial brightness under rapidly changing lighting conditions during driving.
	To address this gap, we propose a highly-adaptive exposure controlling framework that proactively adjusts exposure parameters based on predictive modeling of historical skin reflections. Unlike standard auto-exposure, our method is specifically optimized for rPPG measurement, ensuring the skin region of interest (ROI) remains within the optimal dynamic range for rPPG signal extraction.
	As an important contribution of this study, we introduce ExpDrive, a public in-vehicle physiological monitoring dataset comprising synchronized facial video and reference ECG from 48 subjects captured under real driving conditions. 
	Extensive experiments demonstrate that our method consistently outperforms fixed exposure and standard auto-exposure strategies. 
	Specifically, it reduces the Mean Absolute Error (MAE) by 6.31\,bpm (from 14.1 to 7.79\,bpm) and significantly increases the success rate by 32.3 percentage points (p \textless 0.001) (from 24.9\% to 57.2\%) 
	across challenging driving scenarios.
	Notably, it clearly improved the performance of non-contact heart-rate monitoring in both low-light (rainy) and high-glare (sunny) conditions, validating the efficacy of exposure-aware acquisition design. 
	
\end{abstract}


\section{Introduction}
\label{sec:introduction}
\IEEEPARstart{N}{on-contact} in-vehicle health monitoring—particularly real-time heart rate (HR) estimation for drivers—holds significant promise for enhancing both driving safety and personal health management. 
Continuous monitoring of cardiac activities enables the early detection of acute cardiovascular events (e.g. cardiac arrest, stroke), providing timely alerts that can prevent traffic accidents. Furthermore, the analysis of Heart Rate Variability (HRV) offers a reliable metric for assessing driver fatigue and drowsiness~\cite{Jung2014}. Beyond safety, high-fidelity physiological data recorded during driving can serve as objective forensic evidence in post-accident investigations, clarifying whether a medical emergency (e.g., sudden cardiac arrest) precipitated the incident.

\begin{figure*}[!h]	
	\centerline{\includegraphics[width=1\textwidth]{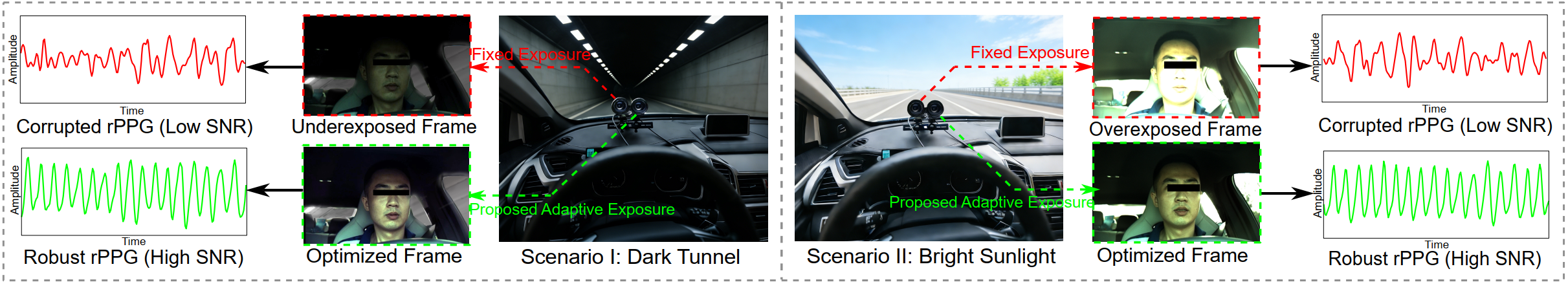}}	
	\caption{
		Conceptual overview of the proposed ``Optimize-at-Capture" framework in dynamic driving environments. The top row (red path) illustrates signal degradation caused by fixed exposure time in tunnel (left) and bright sunlight (right) scenarios, while the bottom row (green path) demonstrates how our adaptive exposure strategy ensures robust rPPG signal extraction.
	}
	\label{fig_1}
\end{figure*}

Driver health monitoring has traditionally relied on contact-based methods such as electrocardiography (ECG) electrodes on steering wheels~\cite{Jung2014,4649406} or seatbelt-integrated sensors~\cite{7991924}. 
While non-invasive, these require sustained, stable mechanical contact between body and sensor for reliable signal acquisition, making them prone to motion artifacts caused by hand movements or clothing friction~\cite{7991924}. Furthermore, such systems often impose unnatural constraints, requiring drivers to maintain specific postures or grip positions to ensure effective sensor coupling~\cite{Jung2014}.
In recent years, remote photoplethysmography (rPPG) has emerged as a compelling non-contact alternative. By analyzing subtle color variations in facial video frames—induced by subcutaneous blood volume changes—rPPG enables contactless extraction of pulse wave signals and subsequent HR estimation~\cite{Huang2023ChallengesAP,wang2024}.


Despite its potential, rPPG faces unique challenges when deployed in real-world driving environments. First, while motion artifacts—common in fitness and general monitoring scenarios—pose a challenge, they can be effectively addressed by existing face tracking algorithms~\cite{wang2024}. In the in-vehicle context, however, the primary bottleneck is the extreme and rapid variation in lighting conditions, as vehicles frequently transit between tunnels, shadows, and direct sunlight—dramatic changes that severely undermine the reliability of rPPG signals~\cite{s18093080,3}. 
To address this, one line of research leverages Near-Infrared (NIR) imaging to mitigate the effects of visible light fluctuations~\cite{1,2,3,4,15}, yet rPPG signals extracted from NIR videos often suffer from the challenge of low signal-to-noise ratios (SNR) due to the low pulsatility issue~\cite{13,14}. 
Other approaches seek to enhance robustness by fusing RGB and NIR videos~\cite{7,8}, or even incorporating millimeter-wave radar data~\cite{physDrive}. 
Although effective, the multi-modal sensor fusion are not ubiquitous in current vehicle cabins.

Consequently, this paper focuses on solutions that utilize ubiquitous RGB cameras for HR monitoring. 
Recent works in this domain focused on data analysis, leveraging complex signal processing or deep learning models to compensate for noise and artifacts after the video has been recorded~\cite{5,6,61}.
However, these methods operate on the premise that the raw video data contains recoverable signal information. In driving scenarios with frequent and intense lighting variations, video acquisition often suffers from over- or under-exposure, leading to irreversible degradation of the underlying RGB signal, as shown in the top row (red path) in Fig.~\ref{fig_1}. Once such information is lost at the acquisition stage, even the most sophisticated post-processing algorithms cannot reconstruct usable physiological data.

To address this limitation, we propose a novel ``optimize-at-capture" framework that integrates on-board real-time video processing with camera parameter controlling at the hardware-level. Specifically, our approach dynamically adjusts the camera’s \textit{exposure time} in real time to maintain the driver’s facial region within an optimal brightness range, thereby preserving rPPG signal integrity at the capturing phase, as shown in the bottom row (green path) in Fig.~\ref{fig_1}. Conventional exposure control strategies~\cite{5952672,Schulz2007,4682088,Kim2014} 
are primarily optimized for visual experience or machine vision tasks and are ill-suited for physiological signal fidelity. While a few exposure control methods~\cite{10781984,10208824} have been proposed for physiological monitoring, they were not tailored for suppressing rapid and unpredictable lighting changes in the in-vehicle scenarios, often exhibiting excessive response latency that compromises the accuracy of physiological measurement.

Furthermore, the field of non-contact in-vehicle physiological monitoring suffers from a scarcity of publicly available datasets, as summarized in Table~\ref{tab:dataset}. Many existing datasets are collected in controlled laboratory conditions or focus on passengers, and—most critically—provide video recordings acquired under a single fixed exposure setting. In real-world driving scenarios, where illumination can vary dramatically, such fixed exposure configurations frequently result in frames that are already overexposed or underexposed at capture time (red path in Fig.~\ref{fig_1}). This severely compromises the fidelity of subsequent rPPG signal extraction, as the essential pulsatile reflectance variations may be irreversibly lost in saturated or disturbance-dominated regions.

In light of these challenges, our contributions are threefold:
\begin{itemize}
	\item Source-aware active exposure adjustment framework: To our knowledge, this is the first method specifically designed to physically optimize the rPPG signal quality at the video capturing stage for in-vehicle settings. By enabling real-time, ROI-guided exposure control during video acquisition, we enhance the rPPG quality at the data capture stage—an approach with broad methodological implications.
	
	\item Adaptive exposure control algorithm: We propose a robust adaptive exposure algorithm based on triplet-frame linear fitting, capable of adjusting the exposure time in real time to compensate abrupt lighting changes. This enables robust rPPG signal extraction even under highly dynamic in-vehicle illumination conditions.
	
	\item Public ExpDrive dataset: We release a new dataset\footnote{The dataset URL will be made publicly available upon acceptance.} captured in real on-road driving scenarios. Crucially, \textit{ExpDrive is the first public dataset that synchronously records multiple videos at different exposure times}.
	

\end{itemize}

\begin{table*}[t]
	\centering
	\caption{Overview of Existing Datasets for Non-Contact In-Vehicle Physiological Monitoring.}
	\label{tab:dataset}
	\small
	\begin{tabular}{lcccccc}
		\toprule
		\textbf{Refs} & \textbf{Scales} & \textbf{Modalities} & \textbf{Driving Scenarios} & \textbf{ \makecell{Exposure \\ Mode}} & \textbf{Availability} \\
		\midrule
		~\cite{dataset1} & 19 drivers; 15 min/driver & ECG, PPG; RGB videos    & Real driving & Single & No \\
		~\cite{dataset2} & 9 drivers; 1506 videos  & HR, RR  & Real driving & Single  & No \\
		~\cite{dataset4} & 64 drivers; 310 clips  & BVP, TEMP, EDA; RGB videos & \textit{Simulated lab} & Single & No \\
		~\cite{10113355} & 10 drivers; 40 min/driver  & PPG, HR; RGB and NIR videos & Real driving & Single & No \\
		~\cite{1} & 7 drivers; 20-60 min/driver  &  PPG; NIR videos & Real driving & Single & No \\
		~\cite{2} & 12 drivers; 3 min/driver  &  PPG; NIR and RGB videos & \textit{Simulated lab} & Single & No \\
		~\cite{61} & 19 drivers or \textit{passengers}; 23 hours   & PPG; RGB videos &\makecell{Real driving;\\ Vehicle Diversity}  & Single & No \\
		~\cite{3} & \textit{18 passengers}; 37 clips   & SpO$_2$; NIR and RGB videos & Real driving & Single & Yes \\
		~\cite{physDrive} & 48 drivers; 30 min/driver  &\makecell{ECG, RR, BVP, SpO$_2$;\\ NIR, RGB, and mmWave} & \makecell{Real driving;\\ Vehicle Diversity}  & Single & Yes \\
		\textbf{Ours} & \textbf{48 drivers; 30-70 min/driver}   & \textbf{ECG, RR; RGB videos} & \makecell{\textbf{Real driving;}\\ \textbf{Vehicle Diversity}}  & \textbf{Multiple} & \textbf{Yes} \\
		\bottomrule 
	\end{tabular}
\end{table*}

\section{Related work}
\label{sec2}
\subsection{Non-Contact Physiological Monitoring in Vehicles}

Traditional HR measurement techniques, such as adhesive ECG patches or finger-clip photoplethysmography (PPG) sensors, are accuracy but require direct skin contact~\cite{10}. In a driving context, such tethered or wearable devices can cause discomfort, restrict movement, and potentially compromise safety~\cite{Jung2014}.
To reduce intrusiveness, unobtrusive contact-based alternatives have been developed, such as capacitive ECG electrodes embedded in steering wheels~\cite{4649406} or seats~\cite{7991924}. While these systems eliminate the need for wires and patches, they still require continuous physical contact between the driver and the sensor.
This reliance significantly undermines their reliability in dynamic driving conditions, where hand movements, clothing friction, and vehicle vibrations introduce substantial motion artifacts. Furthermore, such systems often impose unnatural constraints, requiring drivers to maintain specific postures or grip positions.
In contrast, rPPG has emerged as a compelling non-contact solution. By analyzing subtle skin color variations induced by subcutaneous blood pulsation via video cameras, rPPG enables physiological monitoring without any physical constraints~\cite{Huang2023ChallengesAP,wang2024}. However, robust in-vehicle rPPG implementation faces a critical obstacle: \textit{dynamic illumination}.

To address lighting challenges, NIR imaging has been widely explored~\cite{1,2,3,4,15}. However, NIR-based rPPG measurement faces inherent limitation: rPPG signals extracted from NIR wavelengths typically exhibit a significantly lower SNR compared to the green channel in an RGB video~\cite{13,14}, reducing measurement precision. To balance the robustness to ambient lighting changes and rPPG signal strength, the fusion strategies of combining RGB and NIR streams~\cite{7,8,Hu2024} or video with radar~\cite{physDrive} have been proposed. While fusing complementary data streams can enhance performance, it increases system complexity (e.g., requiring precise sensor calibration and synchronization). Therefore, developing robust algorithms for ubiquitous RGB cameras remains crucial.
Wu et al.~\cite{5} proposed a personalized neural network to mitigate luminance variations and later introduced a CNN-based error compensation model to denoise signals~\cite{6}. Other approaches include hybrid statistical-Monte Carlo frameworks~\cite{61}, signal decomposition~\cite{2,6}, and quality-guided spectrum screening~\cite{5}.

However, a fundamental flaw remains in all aforementioned methods: they operate exclusively after data acquisition. In driving scenarios, rapid lighting transitions (e.g., tunnels, tree shadows) frequently cause standard cameras to over-expose or under-expose. Once the facial ROI is saturated or too dark, the physiological information is irreversibly lost at the sensing level. If physiological data cannot be physically obtained by the camera, no post-processing can recover data that was not successfully captured.

\subsection{Exposure Control Strategies}

In camera-based physiological monitoring within vehicular scenarios, real-time exposure control is crucial for preventing over- and under-exposure, thereby enabling more accurate extraction of physiological information. As noted by Odinaev et al., adjusting exposure time is more effective than Gain for maintaining the rPPG signal integrity~\cite{10208824}. However, existing auto-exposure (AE) methods—such as those based on zonal quality assessment~\cite{5952672}, histograms~\cite{Schulz2007}, mean-median luminance difference~\cite{4682088}, or Gaussian sampling~\cite{Kim2014}—are primarily designed for human or machine vision tasks, and are not optimized for physiological signal fidelity.

A few studies have attempted to optimize exposure specifically for rPPG. Yi et al.~\cite{10781984} applied a Proportional-Integral-Derivative (PID) controller~\cite{1453566} to adjust shutter speed. While theoretically sound, PID controllers in this context often suffer from slow convergence and oscillation, making them unable to keep up with the high-frequency lighting changes of a fast moving vehicle. Laurie et al.~\cite{Laurie2020} proposed to maximize the exposure time within the linear range to boost SNR. However, their method requires a long initialization period ($>1$s), which is impractical for real-time driving applications. Similarly, Odinaev et al.~\cite{10208824} suggested using the quality of the extracted PPG signal itself as the feedback to tune exposure. This creates a ``chicken-and-egg'' issue: obtaining an optimal exposure requires a high-quality rPPG signal, yet the generation of a high-quality rPPG signal itself needs an optimal exposure.

\subsection{In-Vehicle Physiological Datasets}
The development of robust in-vehicle monitoring systems has been hampered by a scarcity of public, real-world datasets. As summarized in Table~\ref{tab:dataset}, existing datasets are often limited in scale, diversity, or realism. A key limitation in scale is exemplified by several datasets with small cohorts, such as those by Hernandez-Ortega et al.~\cite{1} (7 subjects), DriverMVT~\cite{dataset2} (9 subjects), and Xu et al.~\cite{10113355} (10 subjects). 
Some datasets, such as MMDE~\cite{dataset4} and MR-NIRP Indoor Dataset~\cite{2}, including multiple modalities but were collected in \textit{simulated} environments. Although the MR-NIRP Car Dataset~\cite{3} was collected inside a vehicle, all participants were seated as stationary passengers rather than actively engaged in driving tasks.

A common limitation across all existing datasets is the use of a fixed exposure setting or default auto-exposure setting during video acquisition. This approach is sufficient for evaluating post-processing algorithms but is fundamentally inadequate for developing or validating acquisition-level solutions tailored to dynamic in-vehicle environments. 

\section{The proposed method }
\label{sec:3}
The proposed framework introduces a hardware-software iterative algorithm, termed the ``Optimize-at-Capture" strategy. This section details the theoretical derivation of the exposure-skin brightness relationship for optimizing the rPPG measurement, the adaptive control algorithm, and the subsequent signal processing pipeline.

\subsection{Modeling the Relationship Between Exposure Time and Skin-pixel Intensity}
\label{sec3.1}
Let \( L \) denote the incident illuminance on the skin surface, \( R(t) \) the time-varying skin reflectance (which encodes subtle pulsatile variations due to blood volume changes), and \( K \) the camera sensor’s responsivity constant. The total luminous energy \( E \) accumulated by the sensor over the exposure time \( T \) is modeled as~\cite{pos}:

\begin{equation}
	E = K \int_{0}^{T} L \cdot R(t) \, dt.
\end{equation}

This formulation reflects the fact that image sensors integrate incoming photons over the exposure interval rather than sampling instantaneously. The resulting pixel intensity \( I \) is proportional to this accumulated energy, i.e., \( I \propto E \).

The skin reflectance \( R(t) \) can be decomposed into a dominant DC component and a much smaller AC component:

\begin{equation}
	R(t) = R_0 + \Delta R(t),
\end{equation}
where \( R_0 \) is the static diffuse reflectance (DC component), and \( \Delta R(t) \) represents the time-varying component (AC component) due to blood pulsation, which is the rPPG signal itself, with \( |\Delta R(t)| \ll R_0 \).

Substituting Eq. (2) into Eq. (1) yields:

\begin{equation}
	E = K L R_0 T + K L \int_{0}^{T} \Delta R(t) \, dt.
\end{equation}

Accordingly, the observed pixel intensity can be expressed as the sum of DC and AC contributions:

\begin{equation}
	I(T) = I_{\text{DC}}(T) + I_{\text{AC}}(T),
\end{equation}
where \( I_{\text{DC}}(T) \propto K L R_0 T \), and \( I_{\text{AC}}(T) \propto K L \int_{0}^{T} \Delta R(t) \, dt \).

From this model, we have two fundamental observations:

Observation 1: The DC component \( I_{\text{DC}}(T)\) is \textit{linearly proportional} to the exposure time \( T \). This linearity is the basis of our adaptive exposure control strategy. 

Observation 2: The AC component  \( I_{\text{AC}}(T)\) is the temporal integral of \( \Delta R(t) \) over the exposure \( T \). In theory, if \( T \) is comparable to or longer than the cardiac cycle duration ($\sim$ 1 s), this integration acts as a low-pass filter, attenuating high-frequency components and distorting the rPPG waveform. However, in practical in-vehicle deployments, exposure times are typically constrained to less than 30\,ms—significantly shorter than the period of one heart pulse. Therefore, the integral simplifies, and the AC component also exhibits an approximate linear relationship with \( T \):

\begin{equation}
	I_{\text{AC}}(T) \propto K L \Delta R(t_0) T.
\end{equation}

\begin{figure}[!h]	
	\centerline{\includegraphics[width=0.4\textwidth]{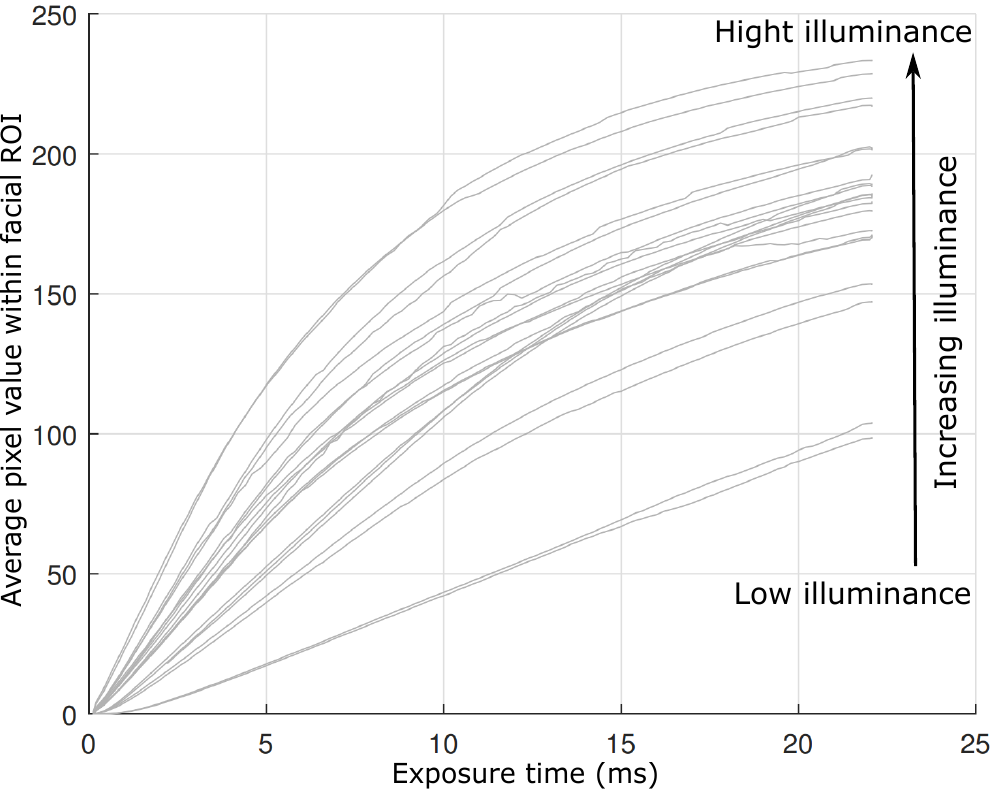}}	
	\caption{
		Characteristic curves showing the relationship between exposure time and average ROI pixel intensity across varying outdoor sunlight conditions. The curves range from low illuminance (bottom, darker scenes) to high illuminance (top, brighter scenes), illustrating how pixel intensity saturates faster under natural stronger light. 
	}
	\label{fig_exp-pxl}
\end{figure}

In an ideal sensor with an unbounded pixel value range, pixel intensity would maintain this linear relationship with exposure time indefinitely. However, the existing camera systems have a finite bit depth (e.g., 255 for 8-bit depth). If $I(T) > I_{max}$, the signal is clipped (saturated), causing the observed relationship to become sub-linear. Fig.~\ref{fig_exp-pxl}, which plots the relationship between exposure time and average pixel value within facial ROI under different illumination intensities, empirically validates this response.
Consequently, the optimal exposure control problem is formulated as a constrained maximization:
\begin{equation}
	T_{opt} = \max { T } \quad \text{s.t.} \quad \mu_{ROI}(T) \leq I_{target} < I_{sat},
\end{equation}
where $\mu_{ROI}(T)$ is the mean intensity of the facial region of interest (ROI), and $I_{target}$ is a safety threshold chosen to prevent saturation caused by local specular reflections.

\subsection{Robust Adaptive Exposure Control Framework}
\subsubsection{Fundamental Assumption}

\begin{figure}[!h]	
	\centerline{\includegraphics[width=0.5\textwidth]{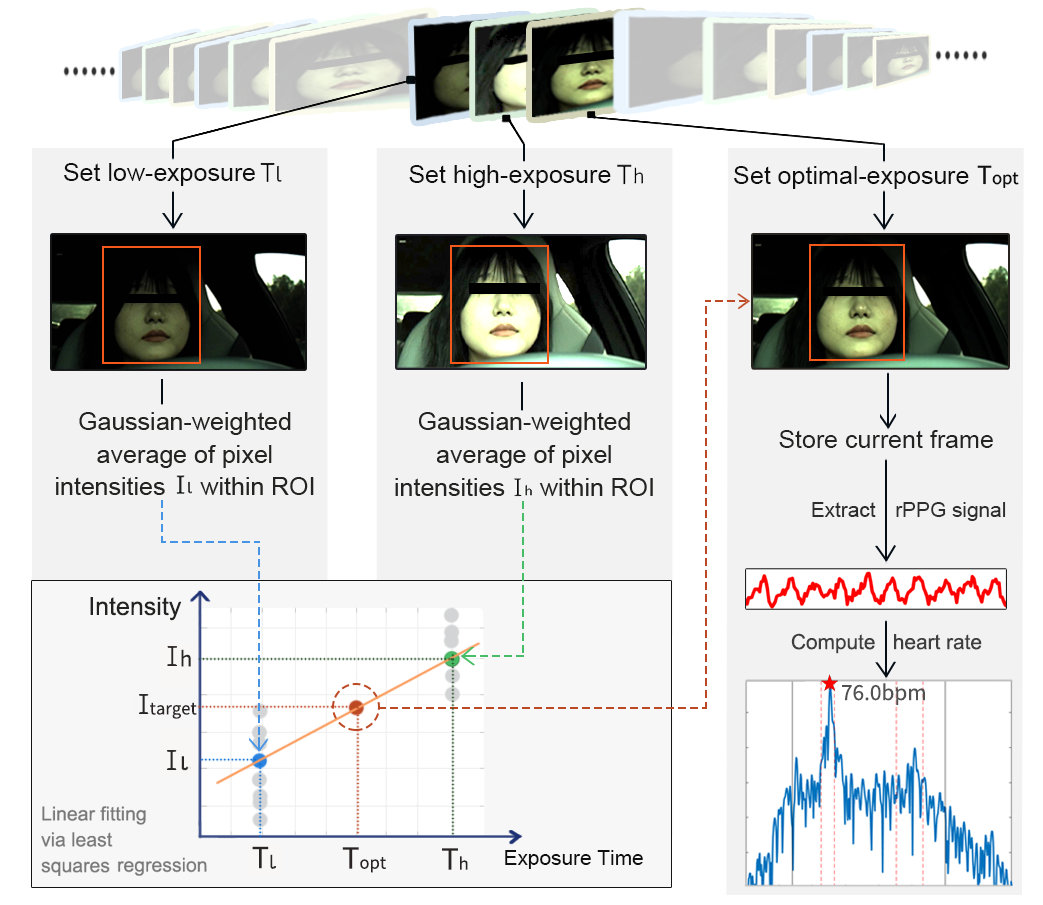}}	
	\caption{
		Triplet-frame adaptive exposure control flowchart.
	}
	\label{fig_3frames}
\end{figure}

Accurately identifying the precise transition point between the linear and nonlinear response regions of a camera sensor—i.e., the onset of pixel saturation—requires detailed analysis of the full intensity distribution within the facial ROI. However, such fine-grained characterization is computationally intensive (typically exceeding 1 second per estimation)~\cite{Laurie2020}, rendering it unsuitable for real-time exposure adaptation in dynamic driving environments. To address this limitation, we restrict our operating regime to the linear portion of the sensor response and propose an efficient triplet-frame adaptive exposure control scheme, illustrated in Fig~\ref{fig_3frames}.

The core premise is that within a short temporal window and a stable scene, the relationship between exposure time \( T \) and the average facial pixel intensity \( I \) remains linear: 
\begin{equation}
	I = k \cdot T + b. 
\end{equation}

Our algorithm dynamically estimates the parameters \( k \) (slope, reflecting the influence of exposure on brightness) and \( b \) (intercept) to adapt to the changing illumination.

\subsubsection{Algorithmic Procedure}

The camera operates at a constant frame rate of 45\,fps. Within each triplet-frame cycle, while the acquisition frequency remains fixed, the exposure time of each individual frame is independently modulated to achieve the desired luminance levels. Each cycle is executed as follows:


Low-Exposure Sample (Frame 1): Acquired with a short exposure time \( T_l \), yielding a low-exposure image from which the mean facial intensity \( I_l \) is computed.

High-Exposure Sample (Frame 2): Acquired with a longer exposure time \( T_h \), producing a high-exposure image with corresponding mean intensity \( I_h \). 

Note that the values of \( T_l \) and \( T_h \) are dynamically adjusted based on the lighting intensity to ensure they remain within the linear range and avoid saturation. In our experimental setup, $T_l$ is typically configured within the range of $5$–$10$\,ms, while $T_h$ is set between $15$–$22$\,ms.

Optimal-Exposure Capture (Frame 3): The two sample pairs obtained at Frame 1 and 2,  \( (T_l, I_l) \) and \( (T_h, I_h) \), are used to fit the linear model of Eq. (7). The parameters can be calculated directly as:
\begin{equation}
	k = \frac{I_h - I_l}{T_h - T_l}, \quad b = I_l - k \cdot T_l.
\end{equation}
Given a pre-defined \textit{target intensity} \( I_{target} \) (set to 140 in this work, determined via pre-experimental analysis as the value just below widespread saturation), the optimal exposure time \( T_{opt} \) is calculated by inverting the linear model:
\begin{equation}
	T_{opt} = \frac{I_{target} - b}{k}.
\end{equation}
The final exposure time is constrained to the camera's physical limits: \( T_{opt} \in [T_{min}, T_{max}] \). The camera is then configured with \( T_{opt} \) to capture the third frame, which is stored for subsequent rPPG analysis.

\subsubsection{Robust Fitting Using Historical Data}

To mitigate the impact of noise from any single sampling pair, the algorithm can be extended to incorporate historical data. Instead of using only the two most recent points, it maintains a rolling buffer of the latest \( n \) low-high exposure pairs \( \{(T_i, I_i)\}_{i=1}^n \).

The linear parameters \( k \) and \( b \) are then estimated by solving the least-squares regression problem:
\begin{equation}
	\min_{k, b} \sum_{i=1}^{n} (I_i - (k \cdot T_i + b))^2.
\end{equation}
The analytical solution is:
\begin{equation}
	k = \frac{n \sum (T_i I_i) - \sum T_i \sum I_i}{n \sum T_i^2 - (\sum T_i)^2}, \quad b = \frac{\sum I_i - k \sum T_i}{n}.
\end{equation}
Note that when \( n=2 \), this method is equivalent to the direct two-point fitting described above. Using \( n>2 \) enhances robustness by smoothing out transient noise.

\subsubsection{Multi-Exposure Region Fusion (MERF)}
\label{merf}
While global exposure optimization ensures that the overall facial brightness remains within the linear regime, it may still fail to prevent local over- or under-exposure in scenes with strong spatial illumination gradients (e.g., direct sunlight illumination on one side of the face and shadow on the other). To mitigate this, we perform a \textit{multi-exposure fusion} step using the three frames captured in one cycle.

The fused image $ I_{\text{fused}}(x,y) $ is then synthesized as:
\begin{equation}
	I_{\text{fused}}(x,y) = W_l \cdot I_l(x,y)+ W_h \cdot I_h(x,y) + W_{\text{opt}} \cdot I_{\text{opt}}(x,y) ,
\end{equation}
where the pixel-wise weights are determined based on the intensity of the optimally exposed frame $ I_{\text{opt}} $. Specifically, we define two intensity thresholds, $ \tau_{low} = 30 $ and $ \tau_{high} = 220 $. 
For underexposed regions where $ I_{\text{opt}}(x,y) < \tau_{low} $, the long-exposure weight $ W_h $ dominates (e.g., is set to 1) to recover the pulsatile information that would otherwise be buried in sensor noise due to insufficient illumination (low SNR). 
For saturated regions where $ I_{\text{opt}}(x,y) > \tau_{high} $, the short-exposure weight $ W_l $ dominates to prevent clipping caused by pixel saturation. For pixels within the optimal range $ [\tau_{low}, \tau_{high}] $, $ W_{\text{opt}} $ is set to 1.

\begin{figure*}[!h]	
	\centerline{\includegraphics[width=1\textwidth]{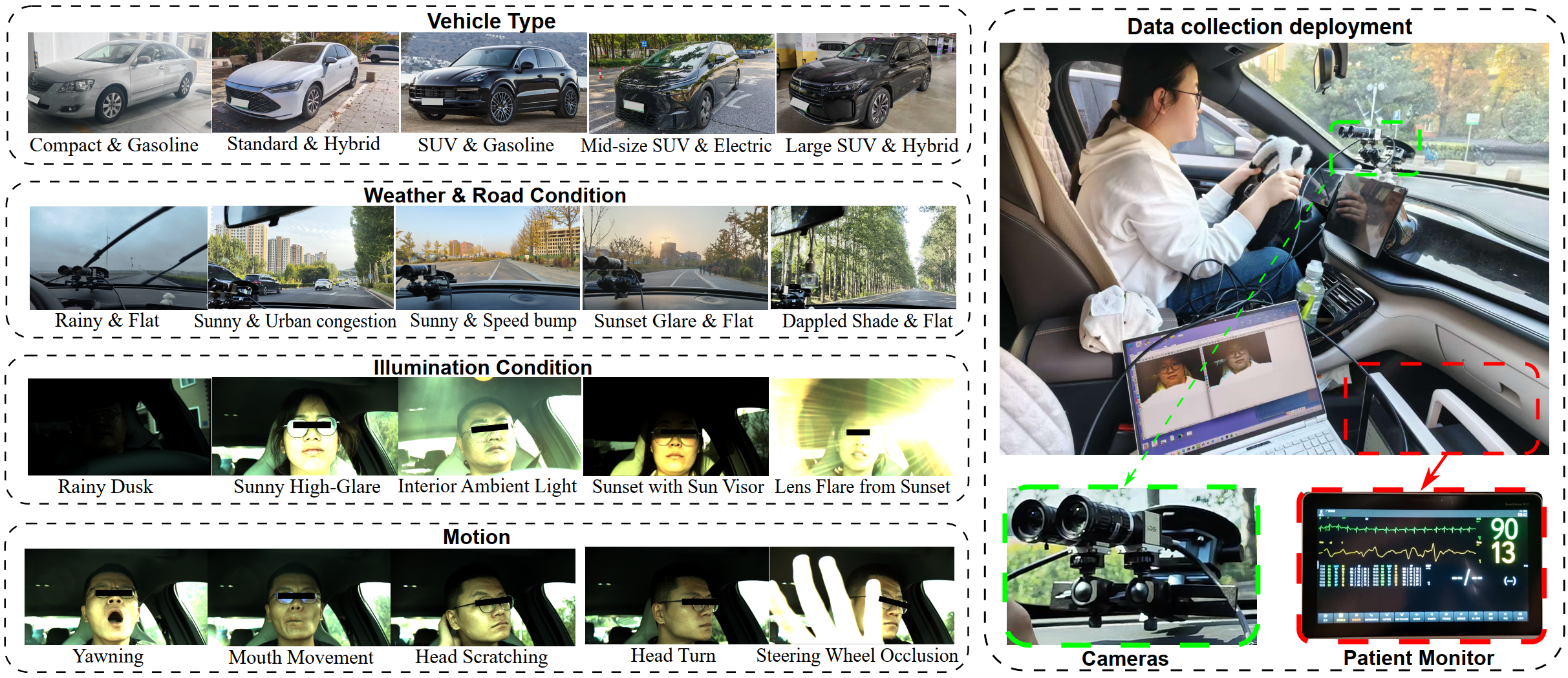}}	
	\caption{
		Overview of the experimental design and in-vehicle setup.
	}
	\label{fig_dataset}
\end{figure*}

\subsection{Physiological Signal Extraction Pipeline}

The system operates continuously by iterating through this triplet-frame cycle. After each cycle, the historical data buffer is updated, discarding the oldest pair if necessary to maintain a size of \( n \), ensuring the linear model to adapt to changing lighting conditions. 
The third frame of each cycle (captured with \( T_{opt} \) and potentially fused) is stored separately, forming a continuous video stream with stable global and local brightness, which is subsequently used for rPPG signal extraction.
The rPPG processing pipeline is as follows:

1.  ROI detection and skin segmentation: The facial ROI is divided into multiple local skin patches, and the average RGB values obtained per patch are concatenated temporally to generate the RGB signals for each patch in the time domian.

2.  Sliding Window Processing: The RGB traces are processed using a sliding window approach. Within each window, signals are normalized and band-pass filtered.

3.  Pulse Waveform Extraction: A core rPPG extraction algorithm (e.g., POS~\cite{pos}) is applied to convert RGB signals into a single-dimensional pulse signal.

4.  Signal Fusion \& Splicing: The top-$K$ patches with the highest SNR are selected and averaged to produce a robust pulse wave for each window. Successive windows are spliced using an overlap-add method to form a continuous waveform.

5.  HR Calculation: The final pulse wave is transformed into the frequency domain via Fast Fourier Transform (FFT). HR is estimated as the frequency of the pulse spectrum peak within [0.6, 3] Hz.  

\section{Experimental setup}
\label{sec4}
\subsection{Dataset: ExpDrive}

As previously noted, publicly available datasets for non-contact in-vehicle physiological monitoring are scarce. To address this gap, we have collected and publicly released a new dataset, named ExpDrive. The data collection was meticulously designed to encompass a wide range of scenarios, as illustrated in Fig.~\ref{fig_dataset}.

\subsubsection{Experimental Design}

To comprehensively investigate the impact of dynamic illumination and driver movement on the accuracy of physiologically extracted signals from video data, the dataset was created under the consideration of four critical dimensions:

\textbf{Weather conditions}: Data were collected under diverse conditions, including bright noon sunshine, overcast/rainy days, direct sunset glare, and dappled shade from roadside trees, to challenge the robustness of rPPG algorithms against intense and fluctuating lighting.

\textbf{Driver Activities}: Participants were instructed to behave naturally, including conversing, yawning, interacting with vehicle controls, and periodically checking rear-view and side mirrors. These actions intentionally introduce realistic facial occlusions, pose changes, and non-rigid motion that challenge signal stability.

\textbf{Road Types}: Three representative road types were selected: smooth highways (minimal vibration), bumpy urban roads (inducing significant whole-body motion), and city commute routes (featuring frequent stops, traffic congestion, and variable ambient light from surrounding infrastructure).

\textbf{Vehicle Types}: To account for differences in driver-to-camera distance, field-of-view, and cabin reflectance properties, data were collected across five production vehicles spanning compact sedans to large SUVs, categorized by Chinese wheelbase standards (Segments A–C) and further annotated by powertrain type: gasoline, battery electric, and hybrid. This ensures generalizability across common vehicle classes in modern fleets.

\subsubsection{Data Collection Protocol}

\begin{figure}[!h]	
	\centerline{\includegraphics[width=0.5\textwidth]{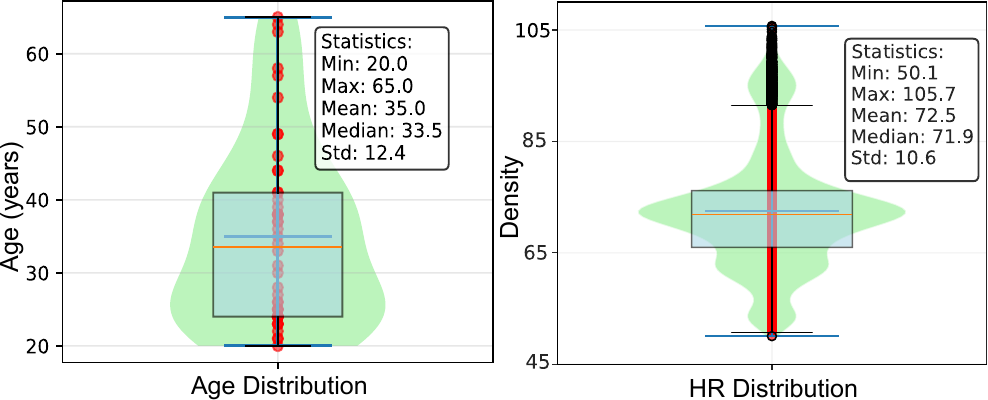}}	
	\caption{
		Age distribution (left) and heart rate distribution (right) of subjects in the ExpDrive dataset. 
	}
	\label{fig_distribution}
\end{figure}

The data collection was conducted simultaneously in two different places, Shenzhen, Guangdong Province, and Qingdao, Shandong Province, China, to ensure geographic and environmental diversity. We recruited 48 licensed drivers (33 male, 15 female), aged 20 to 65 years (mean = 35, SD = 12.4). Their heart rate distribution was approximately normal, as illustrated in Fig.~\ref{fig_distribution}. All participants held a valid driver's license for over one year. All participants were fully informed of the nature and purpose of the study and provided written consent prior to the experiment. This study was approved by the Internal Review Board (IRB) of Southern University of Science and Technology (IRB no.: 20240150).

\subsubsection{Acquisition Hardware}
~\label{hardware}
Two industrial RGB cameras (IDS UI-3160CP) were mounted on the frontal windshield to capture uncompressed video at a resolution of 640 × 480 pixels and a frame rate of 15\,fps. This frame rate was chosen to allow sufficiently long exposure durations in varying light conditions. 
Camera 1 was configured with its manufacture default auto-exposure mode, while Camera 2 implemented our triplet-frame cycle exposure control strategy described in Sec.~\ref{sec:3}. 
By saving each constituent frame from Camera 2 as a separate stream, we obtained distinct video sequences for fixed short exposure, fixed long exposure, and the adaptive exposure output. This synchronized multi-stream acquisition enables a direct, temporally aligned assessment of how different exposure strategies influence raw image quality and the rPPG performance.
Synchronized electrocardiogram (ECG) signals were recorded using a clinical-grade patient monitor (Mindray BeneVision N17) at a sampling rate of 500 Hz. We opted for a chest-lead ECG configuration (using adhesive electrodes) rather than finger-clip PPG as the ground-truth modality because pilot testing revealed that finger-based PPG is highly susceptible to motion artifacts during steering maneuvers—particularly when the driver rotates the steering wheel—leading to unreliable reference signals.
Both video and ECG data were tagged with precise timestamps for temporal synchronization.

\subsection{Quality Metrics}

The quality of the rPPG signals obtained by different exposure time adjustment strategies are evaluated by three metrics.

\textbf{Mean Absolute Error (MAE)} quantifies the average absolute deviation between the estimated heart rate (HR) and the reference HR value. 

\textbf{Success Rate (SR) / Coverage} measures the proportion of time for which the absolute difference between the estimated and reference HR trajectories is within a predefined tolerance \( T \)~\cite{Wang2020DiscriminativeSF}. We adopted a tolerance of \( T = 5 \)\,bpm, a common threshold in the driving scenarios~\cite{Chen24}. It is defined as the percentage of estimates where \( | H\!R_{est} - H\!R_{ref} | \leq 5 \)\,bpm.

\textbf{Signal-to-Noise Ratio (SNR)} evaluates the quality of the extracted rPPG waveform. It is computed by analyzing the power spectral density (PSD) of the pulse signal. 
The SNR (in dB) is defined as the ratio of the signal power within a narrow band centered around the ground-truth heart rate (derived from the synchronized ECG reference) and its first harmonic, to the noise power in the remaining frequency spectrum (typically within [0.7, 4] Hz).

To further evaluate the robustness of the proposed method, we employ the Cumulative Distribution Function (CDF)~\cite{cdf} to visualize the performance variability. For the MAE, where lower values indicate better performance, we employ the standard CDF, defined as $F_X(x) = P(X \le x)$. This represents the probability that the estimation error is within a specific threshold $x$. For metrics where higher values are desirable, such as SNR and SR, we use the complementary CDF, defined as $P(X \ge x)$, which characterizes the probability of achieving a performance level at least as high as $x$. 

\section{Results and discussion}

\begin{table*}[t]
	\centering
	\caption{Overall Performance Comparison of Exposure Control Strategies (N=48)}
	\label{tab:performance_comparison}
	\small
	\begin{tabular}{@{}lcccccc@{}}
		\toprule
		& \multicolumn{2}{c}{\textbf{MAE (bpm)}} & \multicolumn{2}{c}{\textbf{SNR (dB)}} & \multicolumn{2}{c}{\textbf{Success Rate (\%)}} \\
		\cmidrule(lr){2-3} \cmidrule(lr){4-5} \cmidrule(lr){6-7}
		\textbf{Exposure Strategy} & \textbf{Mean ± SD} & \textbf{Imp.} & \textbf{Mean ± SD} & \textbf{Imp.} & \textbf{Mean ± SD} & \textbf{Imp.} \\
		\midrule
		Auto-Exposure & 14.10 ± 5.34 & \textbf{+44.8\%} & -5.12 ± 2.45 & \textbf{+84.0\%} & 24.9 ± 17.2 & \textbf{+129.9\%} \\
		Fixed Short & 10.02 ± 4.15 & \textbf{+22.3\%} & -1.68 ± 3.10 & \textbf{+51.4\%} & 49.4 ± 24.1 & \textbf{+15.7\%} \\
		Fixed Long & 12.47 ± 5.52 & \textbf{+37.5\%} & -3.58 ± 3.01 & \textbf{+77.1\%} & 35.4 ± 22.6 & \textbf{+61.5\%} \\
		\textbf{Proposed: Adaptive} & \textbf{7.79 ± 3.97} & \textbf{--} & \textbf{-0.82 ± 3.43} & \textbf{--} & \textbf{57.2 ± 22.4} & \textbf{--} \\
		\bottomrule
	\end{tabular}
\end{table*}

This section evaluates the performance of different exposure control strategies for in-vehicle HR monitoring. As previously noted, existing in-vehicle camera-based physiological monitoring approaches either rely on a \textit{fixed exposure time} or default to the camera’s built-in \textit{auto-exposure mode}, both of which are ill-suited for the highly dynamic lighting conditions encountered in automotive environments. By comparing our proposed adaptive method against these baseline strategies, we aim to examine how exposure control directly impacts raw image quality and the subsequent accuracy of HR estimation.

\subsection{Overall Performance Evaluation Across All Subjects}

\begin{figure*}[!h]	
	\centerline{\includegraphics[width=1\textwidth]{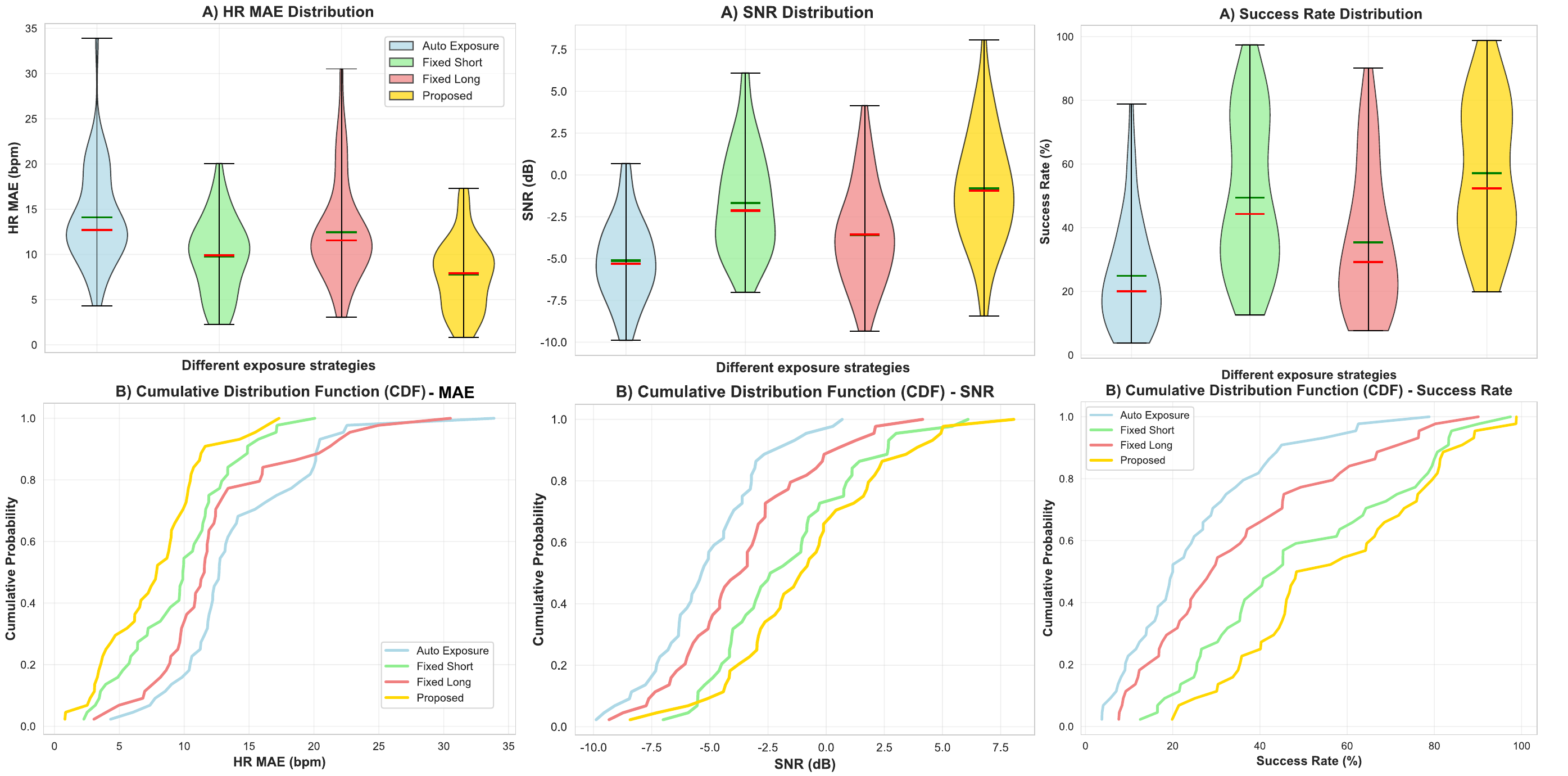}}
	\caption{	
		Distributional comparison of rPPG performance across exposure strategies ($N = 48$). A) Violin plots showing the mean (green bar), median (red bar), interquartile range, and probability density distribution for MAE (bpm), SNR (dB), and SR (\%). B) Corresponding cumulative distribution curves for MAE, $P(\text{MAE} \le x)$; and for SNR and SR, $P(\text{metric} \ge x)$.
	}
	\label{fig_result}
\end{figure*}
This section presents the comprehensive quantitative analysis of all 48 subjects under four exposure adjustment strategies. 
As summarized in Table~\ref{tab:performance_comparison}, the proposed adaptive exposure control strategy demonstrates superior performance across all three evaluation metrics. Compared to default auto-exposure of the IDS camera, our approach achieves a remarkable 44.8\% reduction in MAE, while also delivering substantial improvements of 22.3\% and 37.5\% over fixed short exposure and fixed long exposure strategies, respectively.
The signal quality enhancement is equally impressive, with SNR improvements of 84.0\%, 51.4\%, and 77.1\% compared to auto-exposure, fixed short, and fixed long strategies. Most notably, the success rate—defined as the percentage of measurements within 5\,bpm of the reference HR—shows dramatic gains: a 129.9\% improvement over auto-exposure, 61.5\% over fixed long exposure, and 15.7\% over fixed short exposure. 
It is observable that Fixed Short setting generally outperforms Fixed Long across the entire cohort. This discrepancy is primarily attributed to the distribution of environmental conditions in our dataset, where 36 out of 48 subjects were recorded under bright sunny conditions where over-exposure is more often. Given that the initial fixed parameters were not weather-adapted during early data collection, the $T_h$ setting (22\,ms) frequently led to widespread pixel saturation (pixel value close or equal to 255) within the facial ROI during high-illuminance periods, resulting in irreversible rPPG signal loss. Conversely, the $T_l$ setting (8\,ms) effectively maintained ROI intensities within the sensor’s linear response region under high-glare scenarios, thereby preserving the pulsatile information.

To assess the statistical significance, we conducted non-parametric tests appropriate for paired, non-normally distributed data. A Friedman test revealed significant differences across all exposure strategies for each metric (all \( p < 0.001 \)). Post-hoc Wilcoxon signed-rank tests with Bonferroni correction confirmed that the improvements of the proposed method over each baseline were highly statistically significant (\( p < 0.001 \) for all pairwise comparisons).

The violin plots (Fig.~\ref{fig_result}A-left) reveal that our proposed method achieves the lowest mean HR MAE (7.79\,bpm) with the most concentrated distribution. Beyond the narrow interquartile range (IQR), the shape of the violin for the proposed method exhibits a pronounced ``bottom-heavy" structure, indicating that the majority of subjects are densely distributed within a high-accuracy zone. In contrast, the Auto-Exposure and Fixed Long violins are notably elongated and wider at higher error values (mean: 14.1\,bpm and 12.47\,bpm, respectively), revealing sparse and inconsistent distributions that reflects frequent failures under intense illumination transients.
For SNR and SR metrics—where higher values indicate superior performance—the proposed method’s violins are distinctly ``top-heavy," especially for SR distribution (Fig.~\ref{fig_result}A-right). This density distribution confirms that our approach not only improves average metrics but also ensures high signal quality across the widest range of participants.

The CDF analysis (Fig.~\ref{fig_result}B) provides additional insights into the probabilistic performance characteristics. 
For MAE, the CDF curve of our method lies farthest to the left, meaning a larger proportion of subjects achieve low estimation errors. For instance, approximately 70\% of sessions achieve MAE$ \le 10$\,bpm with our approach, compared to 54\% for fixed short, 33\% for fixed long and only 17\% for auto-exposure. Conversely, for SNR and SR metrics, the CDF curves for the proposed method shift markedly to the right, signifying that a greater proportion of subjects attained higher signal quality and success rates at any performance level.

\subsection{Analysis Across Scenarios and Demographics}

To understand the methods' performance across different categories of challenges (i.e. environment and population), we further evaluate the framework under varying lighting conditions and age groups.
\subsubsection{Performance Under Varying Illumination Conditions}

\begin{figure}[!h]	
	\centerline{\includegraphics[width=0.5\textwidth]{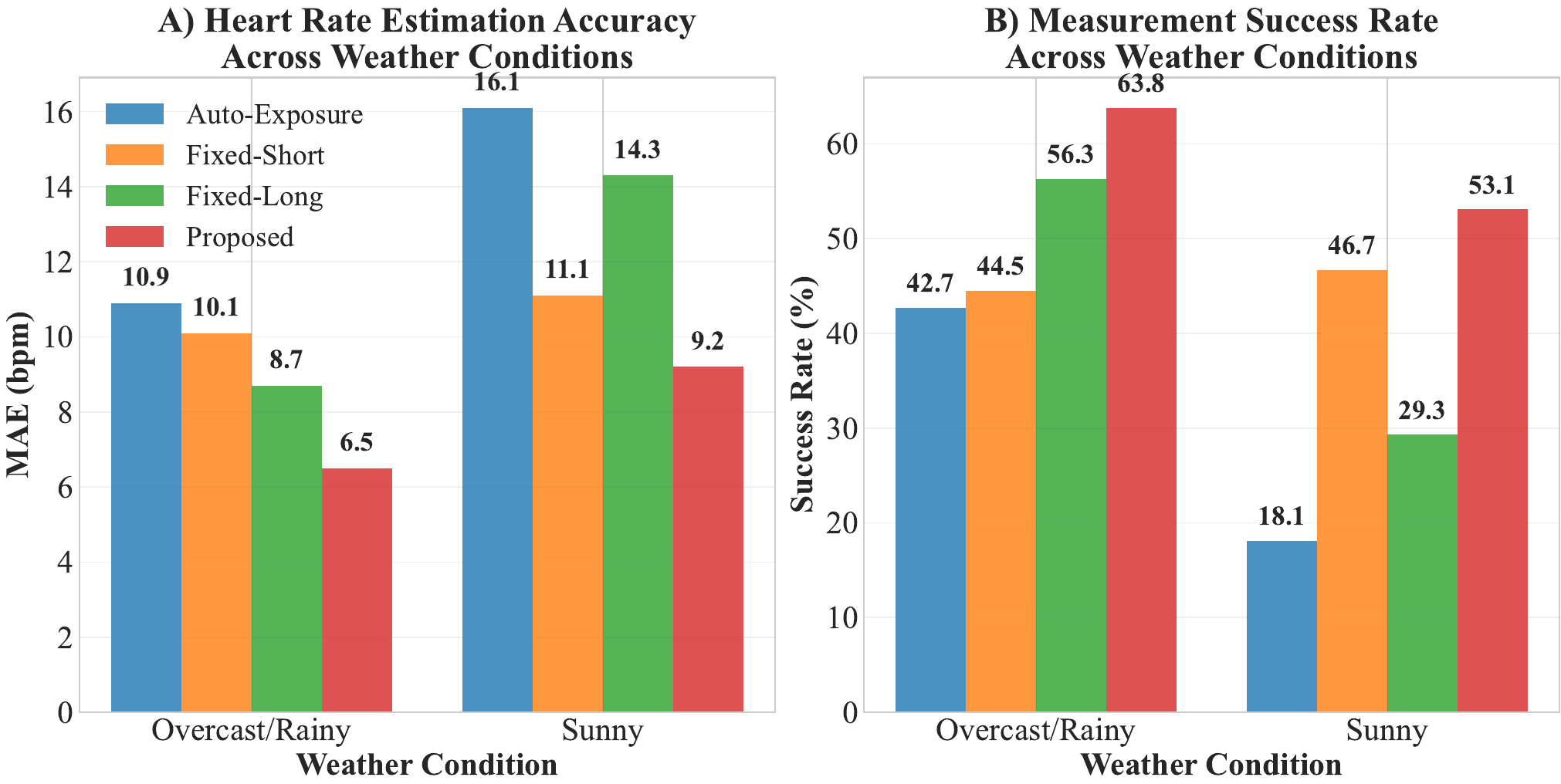}}	
	\caption{
		Performance Comparison Under Different Weather Conditions. 
	}
	\label{fig_weather}
\end{figure}

The influence of ambient illumination conditions was systematically investigated across our dataset, with 12 subjects recorded during overcast/rainy conditions and 36 subjects during bright sunny conditions.
As shown in Fig.~\ref{fig_weather}, auto-exposure performs better in overcast conditions than in sunny ones (MAE: 10.9 vs 16.1\,bpm), as the consistently low illumination reduces the risk of facial region saturation. 
Among fixed-exposure strategies, Fixed Short excels in sunny conditions but underperforms in rain due to underexposure, while Fixed Long shows the opposite trend—validating the intuition that static exposures cannot adapt to scene dynamics. Critically, our adaptive method consistently achieves the lowest MAE and highest success rate across both conditions, demonstrating its robustness to illumination variations. 
Notably, overall performance is better in overcast conditions than in sunny ones. This stems from our initial experimental design: the fixed short \( (T_l) \) and long \( (T_h) \) exposure durations were not weather-adapted during early data collection. Through iterative refinement, we established weather-specific exposure bounds: overcast (\( T_l=8\,ms, T_h=22\,ms \)) and sunny (\( T_l=5\,ms, T_h=16\,ms \)).
This parameter adaptation further enhanced performance, validating the necessity of context-aware exposure calibration.

\subsubsection{Performance Across Age Groups}

\begin{figure}[!h]	
	\centerline{\includegraphics[width=0.5\textwidth]{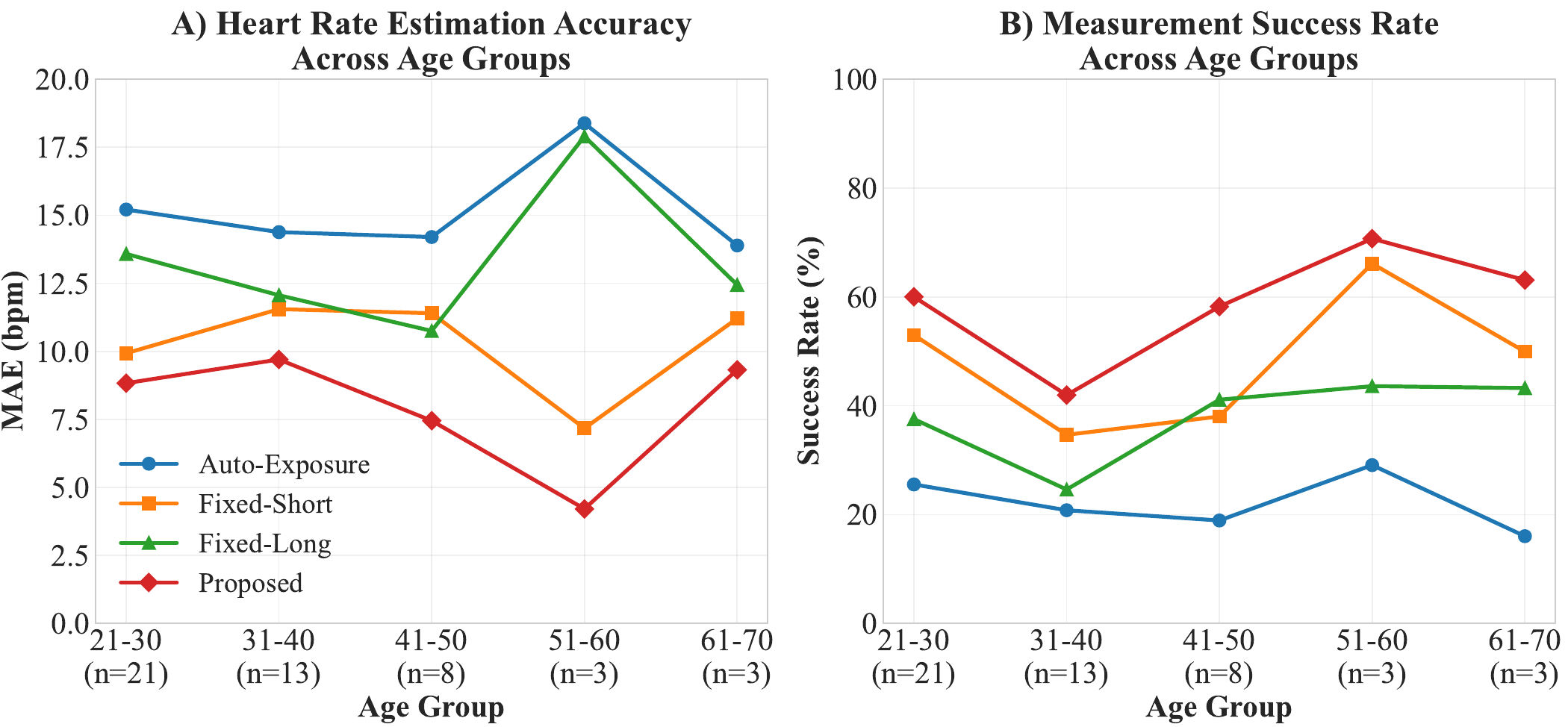}}	
	\caption{
		Comparison of the performance of four exposure control methods across different age groups. 
	}
	\label{fig_age}
\end{figure}

We further analyze the performance in terms of driver age, partitioning the cohort into five groups, as shown in Fig.~\ref{fig_age}. The proposed adaptive exposure strategy consistently outperforms other modes across every age cohort, demonstrating its universal applicability regardless of subject age. While minor performance variations exist between age groups, no statistically significant age-dependent pattern emerges. The 51-60 age group exhibits best performance with our method (MAE: 4.21\,bpm, SR: 70.7\%), though this observation should be interpreted cautiously given the limited sample size (n=3). The consistent superiority across demographics suggests that the fundamental challenge—maintaining optimal facial brightness for rPPG extraction—is independent of age-related physiological factors.

\subsection{Case Studies of Representative Subjects}
\label{case}

\begin{figure*}[!h]	
	\centerline{\includegraphics[width=0.8\textwidth]{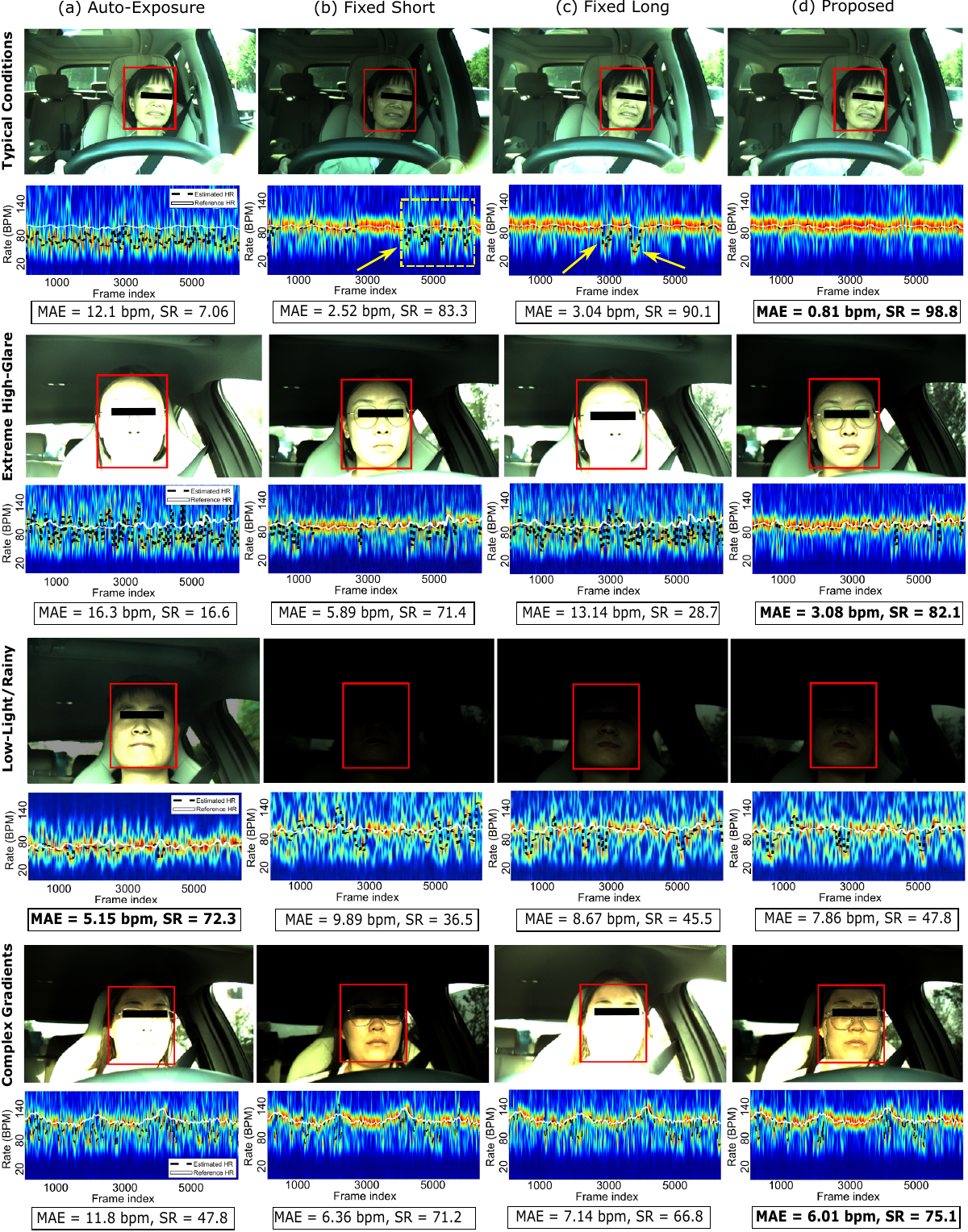}}	
	\caption{
		Qualitative performance comparison across four representative driving conditions. From top to bottom: Case 1 (typical conditions), Case 2 (extreme high-glare), Case 3 (low-light/rainy), and Case 4 (complex local gradients). For each case, we show the sampled video frames (top) and the corresponding time-frequency spectrograms of the extracted rPPG signals (bottom). The results demonstrate the system's superior adaptability in dynamic and high-contrast environments (Cases 1, 2, 4) while acknowledging its hardware-defined exposure limits in near-dark conditions (Case 3).
	}
	\label{fig_spec}
\end{figure*}

To provide a more detailed evaluation of the ``Optimize-at-Capture" framework, we select four representative driving scenarios (illustrated in Fig.~\ref{fig_spec}) for qualitative analysis. These cases encompass a spectrum of illumination challenges, from rapid lighting transitions to extreme edge cases (e.g., severe overexposure caused by direct sunlight and high-contrast spatial gradients induced by sun visor usage).

\subsubsection{Case 1: Typical Driving Conditions with Rapid Lighting Transitions}

The first case (Row 1) represents the most frequent scenarios encountered in daily driving, characterized by moderate ambient light interspersed with rapid shadows from roadside trees and structures. The specific performance of each exposure strategy is sequentially analyzed as follows:

\textit{Default Auto-Exposure (a):} The video frame reveals severe facial overexposure, a consequence of the camera's automatic exposure algorithm attempting to achieve a predetermined mid-gray luminance level for the entire scene. Since dark background regions (e.g., roof and rear seats) dominated the frame composition, the exposure control system incorrectly interpreted the scene as underexposed and consequently increased exposure time, resulting in facial region saturation and consequent rPPG signal loss. 

\textit{Fixed Short Exposure (b):} 
The fixed short exposure strategy produced noticeably underexposed facial regions. The spectral analysis reveals significant estimation errors between frames 4000-6000 (highlighted by the yellow dashed rectangle), coinciding with periods of low ambient illumination combined with rapid shadow transitions from overhanging trees. This combination of inherent underexposure and dynamic lighting variations resulted in substantial signal degradation. 

\textit{Fixed Long Exposure (c):} 
While brighter than the short-exposure variant, this setting suffers from intermittent overexposure, particularly evident in the cheek regions. Spectral analysis identifies prominent estimation errors around frames 3000 and 4000 (indicated by arrows), corresponding to high illumination conditions exacerbated by intermittent shadow patterns. The resulting signal clipping during bright intervals, combined with rapid illumination transitions, compromised measurement accuracy. 

\textit{Proposed Adaptive Exposure (d):} 
In stark contrast, our proposed exposure time adaptation strategy maintained optimal facial brightness throughout the recording session. The video frame exhibits well-balanced exposure without saturation or underexposure issue. Spectral analysis confirms exceptional performance, with a clean, dominant, and well-defined peak at the fundamental heart rate frequency and minimal spectral noise. The algorithm successfully avoided both the overexposure issues observed around frames 3000-4000 and the underexposure problems encountered during frames 4000-6000. Over the entire 7-minute driving, the method achieves outstanding performance: MAE = 0.81\,bpm, SR = 98.8\% (tolerance = 5\,bpm), and SNR = 8.1 dB. 


\subsubsection{Case 2: Extreme High-Glare (Direct Sunlight)}

In scenarios featuring direct sunlight (Fig.~\ref{fig_spec}, Row 2), the ambient illuminance is exceptionally high. The default AE and Fixed Long settings result in total facial saturation. While Fixed Short ($5$\,ms) remains partially viable, localized overexposure still persists on the lower face. Our adaptive framework proactively compresses the exposure time below $5$\,ms, ensuring that no pixels within the ROI exceed the saturation threshold, thereby preserving the underlying pulsatile information.

\begin{figure*}[!h]	
	\centerline{\includegraphics[width=0.9\textwidth]{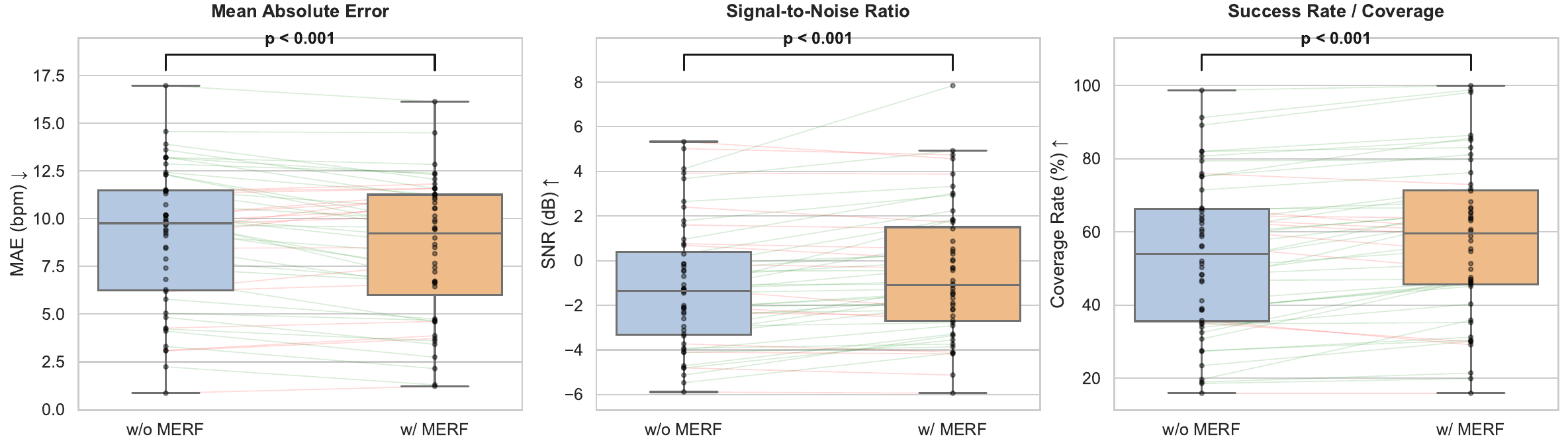}}	
	\caption{
		Performance comparison and ablation study of the Multi-Exposure Region Fusion (MERF) module ($N=48$). The boxplots and paired subject trajectories illustrate the impact of MERF on MAE, SNR, and success rate. Statistical significance is denoted by p-values derived from the Wilcoxon signed-rank test.
	}
	\label{fig_MERF}
\end{figure*}

\subsubsection{Case 3: Low-Light Limitation (Rainy Dusk)}
This case (Fig.~\ref{fig_spec}, Row 3) illustrates the fundamental physical bottleneck inherent in our triplet-frame cycle scheme. To maintain a 15\,fps output for each exposure channel, the hardware must operate at a constant 45\,fps, which imposes a maximum exposure limit of approximately $22.2$\,ms ($1000/45$) per frame. In near-dark environments such as heavy rain at dusk, this ceiling is insufficient to capture an adequate signal, leading to underexposure across all adaptive and fixed channels. Conversely, the camera's default AE, unconstrained by the triplet-frame cycle, can extend exposure to $66.7$\,ms (15\,fps), yielding a brighter image and more accurate HR estimation.

This limitation represents a fundamental trade-off between frame rate and exposure time. It is a physical constraint that any exposure control method must confront: it is inherently impossible to handle extremely dark conditions while maintaining a sufficient frame rate for rPPG extraction. While exposure control-based approach can successfully manage extremely bright cases by shortening the shutter speed, it remains physically constrained in near-dark environments where the required integration time exceeds the available frame interval.

\subsubsection{Case 4: Complex Local Gradients (Sunset with Sun Visor)}
This scenario (Fig.~\ref{fig_spec}, Row 4) presents an extreme spatial luminance gradient: the forehead is in deep shadow under a sun visor, while the lower face is exposed to intense sunset glare. 
In this scenario, both the camera auto-exposure and fixed long exposure ($20$\,ms) resulted in saturation across the majority of the facial ROI, leading to a loss of rPPG information. While fixed short setting maintained reasonable intensity on the lower face, the forehead remained significantly underexposed, providing insufficient SNR in the shaded regions. 

To maintain the global intensity target ($I_{target}$ = 140), the proposed exposure-adaptation algorithm increased the exposure time to brighten the shaded forehead. Although this led to localized saturation in the lower face, our MERF mechanism (Sec.~\ref{merf}) resolved this by substituting pixels exceeding the threshold ($\tau_{high}$ = 220) with their counterparts from the fixed short stream. By combining the optimally exposed forehead from the adaptive frame with the glare-free lower face from the short-exposure frame, the system ensured that the entire facial ROI remained within the linear response range, resulting in a highly accurate heart rate estimation.

\subsection{Ablation Study: Effectiveness of MERF}

Although adaptive exposure control maintains the global average facial brightness within the optimal brightness range, local illumination variations (e.g., strong sidelight or shadows) can still cause regional saturation or underexposure, as shown in Case 4 of Fig.~\ref{fig_spec}. MERF addresses this by fusing three sequentially captured frames ($T_l$, $T_h$, $T_{opt}$) into a single optimized image, preserving details in both highlights and shadows through spatially varying weights. 
We compare the rPPG pipeline using only the globally optimized frame ($T_{opt}$, w/o MERF) against the pipeline using the MERF-fused frame (w/ MERF) on 48 subjects. Fig.~\ref{fig_MERF} presents a paired comparison across three metrics:
1) \textbf{Accuracy}: The MAE decreased from 9.44\,bpm (w/o MERF) to 8.83\,bpm (w/ MERF). Most subjects, especially those under high-contrast lighting, showed reduced error.
2) \textbf{Signal Quality}: The average SNR improved from -0.91 dB to -0.49 dB, indicating better recovery of pulsatile information from otherwise clipped or noisy regions.
3) \textbf{Robustness}: The success rate increased from 51.5\% to 56.1\%, reflecting improved system reliability.
However, performance degradation occurred for a minority of subjects (see red lines in Fig.~\ref{fig_MERF}), primarily due to: 1) Temporal misalignment and ghosting: Sequential capture of the three exposures can cause artifacts under rapid head motion or vibration, distorting subtle skin-color variations.
2) Fusion artifacts at boundaries: Unsmooth transitions in the fusion map may introduce intensity discontinuities within the facial ROI. These spatial-temporal gradients can introduce artificial fluctuations in the extracted color traces, which interfere with the pulse signal and degrade the precision of HR estimation.

In summary, while MERF generally enhances performance in high dynamic range scenes, its benefits can be diminished in the presence of severe motion. Overall, the module provides an additional layer of protection against intra-frame lighting variations by balancing highlights and shadows at the pixel level.


\subsection{Limitations and Future Research Directions}

Despite the demonstrated advantages, several limitations remain for future improvement.
First, the current implementation requires manual calibration of exposure parameters ($T_l, T_h$) for different weather conditions. While this approach yields significant improvements over fixed strategies, a fully autonomous system capable of dynamically adapting these parameters based on real-time environmental assessment would represent a substantial advancement.
Second, the triplet-frame cycle architecture imposes inherent constraints on maximum exposure time. To maintain a minimum frame rate of 15\,fps for reliable rPPG extraction, the complete cycle must operate at 45\,fps (15\,fps $\times$ 3), limiting maximum exposure time to approximately 22\,ms. 
This constraint proves adequate under normal illumination conditions but becomes a critical bottleneck in extremely low-light scenarios (e.g., nighttime driving, heavy rainfall, as demonstrated in Case 3 in Sec.~\ref{case}), where the required integration time for a usable signal exceeds the hardware-defined limit. Since this represents an intrinsic limitation of exposure-time modulation, simply refining the control algorithm cannot resolve the SNR loss in near-darkness. To overcome this physical ceiling, future research should consider complementary strategies such as leveraging the camera's analog gain or integrating active NIR light sources to maintain signal fidelity when the maximum exposure time is reached.
Third, the current exposure control operates at a full-frame level, which can lead to local pixel saturation in high-dynamic-range scenarios (e.g., facial regions near windows). While our proposed local brightness fusion provides a partial mitigation strategy, regional exposure optimization would require pixel-level or region-adaptive exposure control. 
\section{Conclusions}

This paper has presented a comprehensive investigation into adaptive exposure control for robust non-contact heart rate monitoring in moving vehicles. Through systematic analysis and experimental validation, we have demonstrated that conventional exposure strategies—whether fixed exposure times or camera auto-exposure modes—are fundamentally inadequate for the specific requirements of physiological signal extraction in dynamic in-vehicle environments.
Our primary contribution lies in the introduction of a novel hardware-software co-design paradigm that implements real-time, ROI-aware exposure control during video acquisition. The proposed triplet-frame cycle strategy successfully maintains facial brightness within the optimal linear response range, thereby preserving rPPG signal integrity from the outset. This ``optimize-at-capture" approach represents a significant departure from existing methods that rely exclusively on post-processing software solutions.
We validated our approach on ExpDrive, a richly annotated dataset featuring synchronized facial video and clinical-grade ECG from 48 drivers.
The experimental results unequivocally show that under diverse real-driving conditions, our adaptive exposure control scheme exhibits consistent and substantial performance improvements compared to other strategies. 


In conclusion, this work establishes that effective exposure control is not merely an optional enhancement but a fundamental requirement for reliable non-contact physiological monitoring in challenging in-vehicle environments. By addressing this critical aspect at the acquisition stage rather than relying solely on post-processing, our approach enables more accurate and robust HR monitoring, paving the way for enhanced driver safety and well-being management through continuous, unobtrusive physiological assessment during driving activities.


\bibliographystyle{elsarticle-num} 
\bibliography{refs}

\begin{IEEEbiography}[{\includegraphics[width=1in,height=1.25in,clip,keepaspectratio]{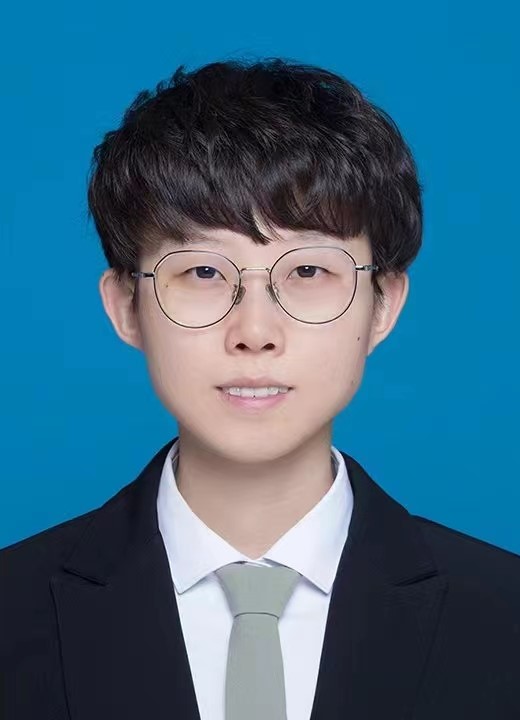}}] {Jieying Wang} received the B.E. degree in electronic and information engineering from China Jiliang University, China, in 2015, and the M.S. in marine information science and engineering from Zhejiang University, China, in 2018, and the Ph.D. degree in computer science and technology from Unviersity of Groningen, the Netherlands, in 2022. She is currently a lecture with Shandong University of Science and Technology, China. Her research interests are image processing and machine learning, with applications in health monitoring.
\end{IEEEbiography}

\begin{IEEEbiography}[{\includegraphics[width=1in,height=1.25in,clip,keepaspectratio]{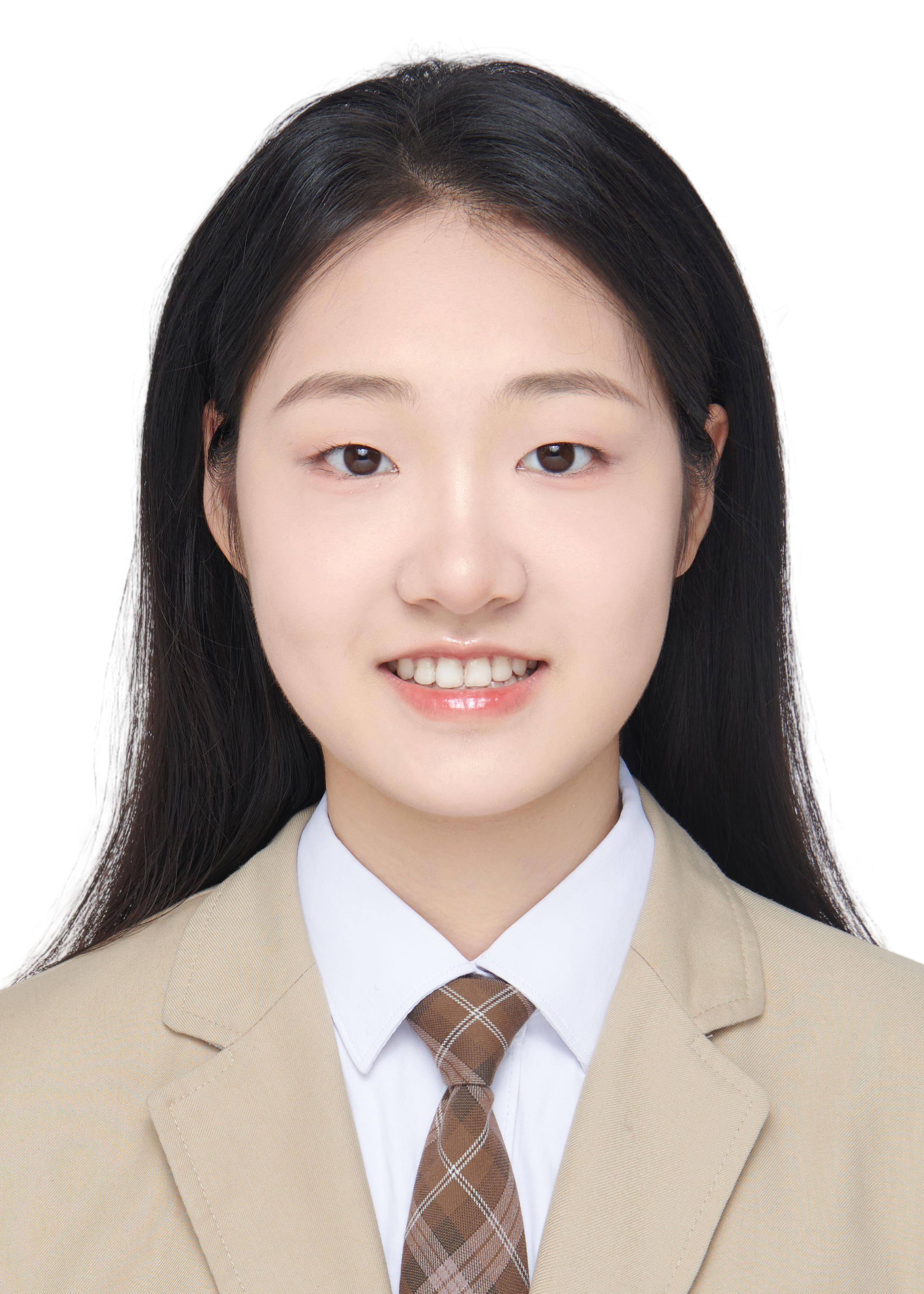}}]{Cai Xinqi} is currently an undergraduate student majoring in Biomedical Engineering at Southern University of Science and Technology (SUSTech), Shenzhen, China. Her research interests include exposure algorithms and health monitoring.
\end{IEEEbiography}

\begin{IEEEbiography}[{\includegraphics[width=1in,height=1.25in,clip,keepaspectratio]{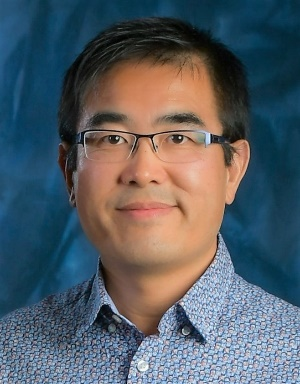}}] {Caifeng Shan} is a full professor with Nanjing University, China. He was previously a Senior Scientist with Philips Research, Eindhoven, The Netherlands. He received the B.Eng. degree from the University of Science and Technology of China, the M.Eng. degree from the Institute of Automation, Chinese Academy of Sciences, and the Ph.D. degree from Queen Mary, University of London. His research interests include computer vision, pattern recognition, medical image analysis, and related applications. He has co-authored about 200 papers and more than 100 patent applications. He has served as Associate Editor for journals including Pattern Recognition, IEEE Journal of Biomedical and Health Informatics, and IEEE Transactions on Circuits and Systems for Video Technology. He is a Senior Member of IEEE.
\end{IEEEbiography}

\begin{IEEEbiography}[{\includegraphics[width=1in,height=1.25in,clip,keepaspectratio]{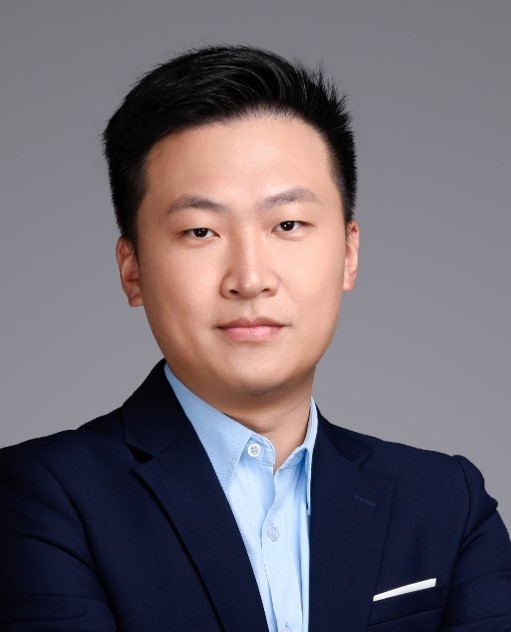}}]
{Wenjin Wang} is an Associate Professor of Southern University of Science and Technology, China. He received the B.Sc. degree from Northeastern University, China (2011), the M.Sc. degree from University of Amsterdam, The Netherlands (2013), and the Ph.D. degree from Eindhoven University of Technology (TU/e), The Netherlands (2017). He was an Assistant Professor of TU/e and a Scientist of Philips Research Eindhoven. He published 130 peer-reviewed scientific papers, 3 academic books, and holds 45 granted patents. He got the Prize Paper Award of IEEE-TBME 2022. He was awarded the National Excellent Young Scholars (Overseas) in 2022. His current research was supported by the National Key R\&D Program of China and Natural Science Foundation of China. 
His research interests include the video health monitoring and its translation into a medical device.
\end{IEEEbiography}

\end{document}